\documentclass[sigconf, screen]{acmart}

\AtBeginDocument{%
  }

\setcopyright{acmlicensed}
\copyrightyear{2018}
\acmYear{2018}
\acmDOI{XXXXXXX.XXXXXXX}

\acmConference[Conference acronym 'XX]{Make sure to enter the correct
  conference title from your rights confirmation email}{June 03--05,
  2018}{Woodstock, NY}

\acmISBN{978-1-4503-XXXX-X/2018/06}

\usepackage{graphicx} 
\usepackage{subcaption}
\usepackage{float}
\begin{document}

\title{HOMER: Homography-Based Efficient Multi-view 3D Object Removal}

\author{Jingcheng Ni}
\authornote{Both authors contributed equally to this research.}
\affiliation{%
  \institution{Duke Kunshan University}
  \city{Kunshan}
  \state{Jiangsu}
  \country{China}
}
\affiliation{%
  \institution{Duke University}
  \city{Durham}
  \state{NC}
  \country{USA}
}
\email{jingcheng.ni@dukekunshan.edu.cn}

\author{Weiguang Zhao}
\authornotemark[1]
\affiliation{%
  \institution{Xi’an Jiaotong-Liverpool University}
  \city{Suzhou}
  \state{Jiangsu}
  \country{China}
}
\affiliation{%
  \institution{University of Liverpool}
  \city{Liverpool}
  \state{}
  \country{UK}
}
\email{weiguang.zhao@liverpool.ac.uk}

\author{Daniel Wang}
\affiliation{%
  \institution{Yale University}
  \city{New Haven}
  \state{CT}
  \country{USA}}

\author{Ziyao Zeng}
\affiliation{%
  \institution{Yale University}
    \city{New Haven}
  \state{CT}
  \country{USA}
}

\author{Chenyu You}
\affiliation{%
 \institution{Stony Brook University}
 \city{Stony Brook}
 \state{NY}
 \country{USA}}

\author{Alex Wong}
\affiliation{%
  \institution{Yale University}
  \city{New Haven}
  \state{CT}
  \country{USA}}

\author{Kaizhu Huang}
\authornote{Corresponding author}
\affiliation{%
  \institution{Duke Kunshan University}
  \city{Kunshan}
  \state{Jiangsu}
  \country{China}}
\email{kaizhu.huang@dukekunshan.edu.cn}

\begin{abstract}
3D object removal is an important sub-task in 3D scene editing, with broad applications in scene understanding, augmented reality, and robotics. However, existing methods struggle to achieve a desirable balance among consistency, usability, and computational efficiency in multi-view settings. These limitations are primarily due to unintuitive user interaction in the source view, inefficient multi-view object mask generation, computationally expensive inpainting procedures, and a lack of applicability across different radiance field representations. To address these challenges, we propose a novel pipeline that improves the quality and efficiency of multi-view object mask generation and inpainting. Our method introduces an intuitive region-based interaction mechanism in the source view and eliminates the need for camera poses or extra model training. Our lightweight HoMM module is employed to achieve high-quality multi-view mask propagation with enhanced efficiency. In the inpainting stage, we further reduce computational costs by performing inpainting only on selected key views and propagating the results to other views via homography-based mapping. Our pipeline is compatible with a variety of radiance field frameworks, including NeRF and 3D Gaussian Splatting, demonstrating improved generalizability and practicality in real-world scenarios. Additionally, we present a new 3D multi-object removal dataset with greater object diversity and viewpoint variation than existing datasets. Experiments on public benchmarks and our proposed dataset show that our method achieves state-of-the-art performance while reducing runtime to one-fifth of that required by leading baselines.
\end{abstract}



\begin{teaserfigure}
  \includegraphics[width=0.96\textwidth]{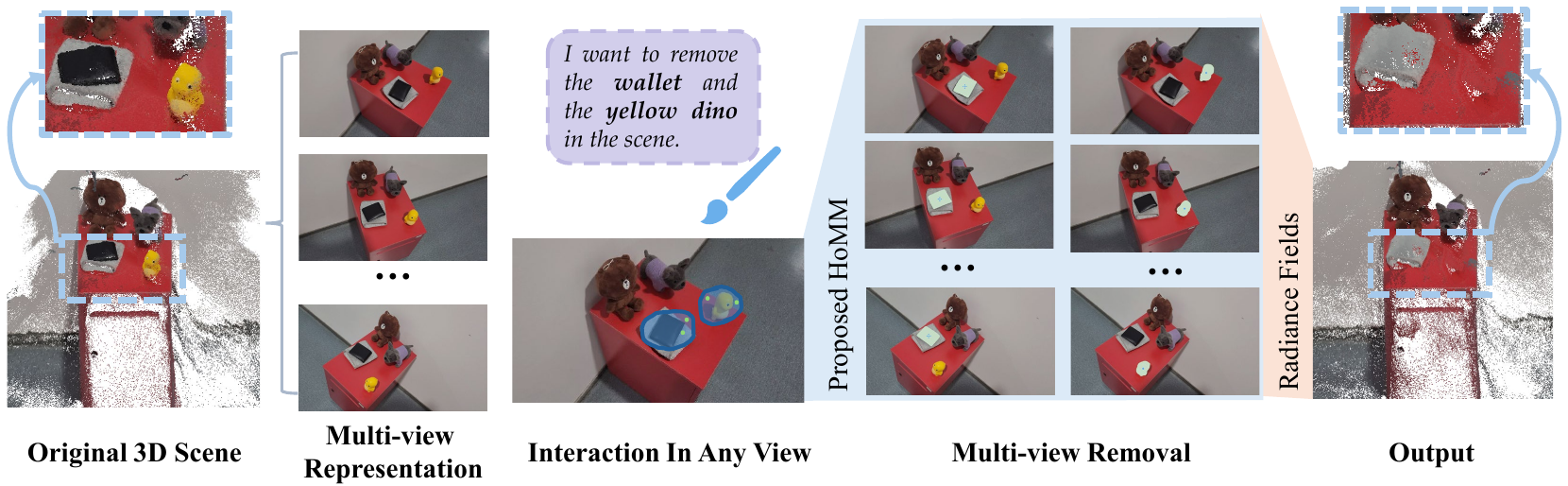}
  \caption{HOMER: Homography-Based Efficient Multi-view 3D Object Removal. The blue shaded area represents the selected removal area. The green dot stands for the selected background and the objects to be retained. HoMM stands for Homography-based Masks Matching.}
  \label{fig:intro}
\end{teaserfigure}

\received{20 February 2007}
\received[revised]{12 March 2009}
\received[accepted]{5 June 2009}

\maketitle

\section{Introduction}

\begin{figure*}[!ht]
 \centering
\includegraphics[width=0.94\textwidth]{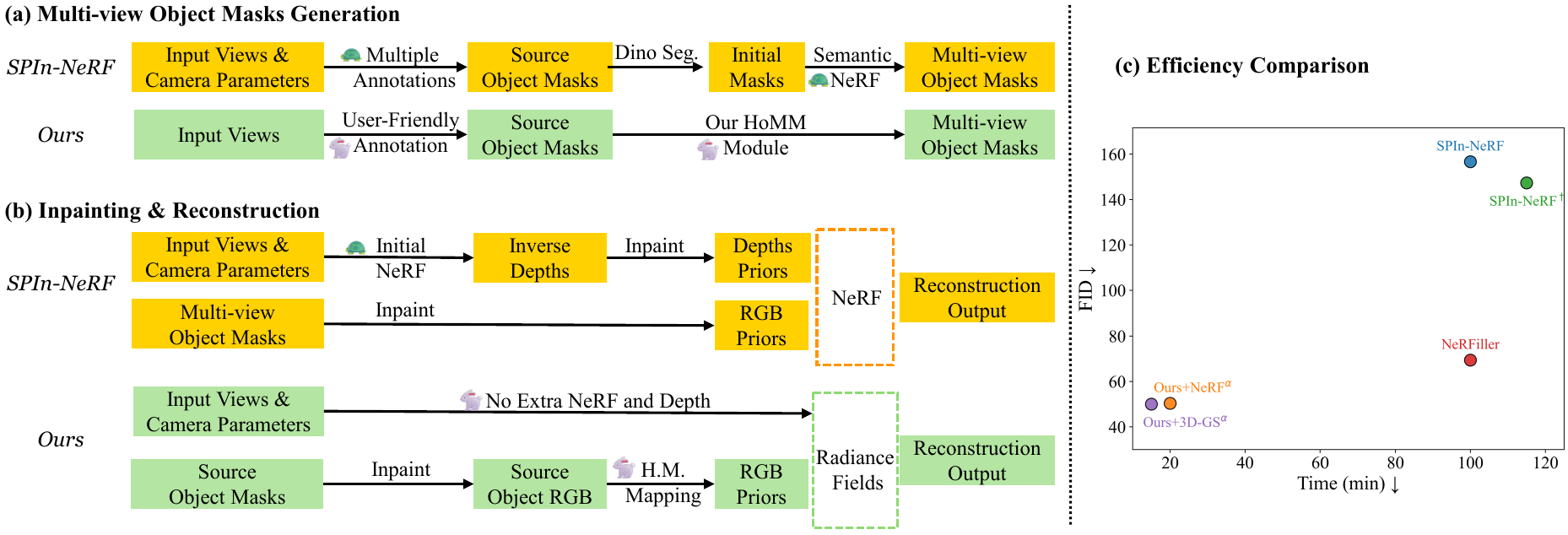} 
 \caption{Comparison of our method with baseline SPIn-NeRF ~\cite{mirzaei2023spin} for multi-view object masks generation (top) and inpainting \& reconstruction (bottom). The turtle icons indicate computationally intensive steps in the SPIn-NeRF pipeline, while rabbit icons highlight the efficiency improvements in our approach.}
 \label{fig:comparision}
 \end{figure*}
 
3D scene editing~\cite{haque2023instruct,fang2024chat,he2024customize} is a fundamental problem in computer vision and graphics, aimed at modifying the structure, content, or appearance of a 3D environment. It plays a critical role in a wide range of applications, including virtual and augmented reality, digital content creation, autonomous systems, and digital twin construction. Among its various sub-tasks, 3D object removal~\cite{mirzaei2023spin,weber2024nerfiller,weder2023removing,mirzaei2023reference} focuses on eliminating specified entities from a 3D scene and reconstructing the missing content in a visually and structurally coherent manner. This task enhances the clarity and usability of 3D data and representations, thereby improving the reliability and accuracy of downstream applications such as object recognition, scene understanding~\cite{zhao2025BFANet}, and robotic perception~\cite{ding2022language}.

Recent 3D object removal~\cite{mirzaei2023spin,weder2023removing,mirzaei2023reference,chen2024mvip,wei2023clutter,wang2024innerf360} has achieved impressive progress, while there still remain challenges, especially in multi-view settings, where the joint requirements of consistency, usability, and computational efficiency are often difficult to satisfy. Two key limitations persist in current pipelines:
1) \textbf{Limitations in multi-view object mask generation.}
First, user interaction in the source view is often inefficient and lacks user-friendliness. 
Although some methods~\cite{mirzaei2023spin,weber2024nerfiller} have recognized the importance of generating consistent object masks across multiple views, they often overlook usability and incur high computational costs. Existing baselines typically require users to a) click on target objects in selected views to guide segmentation using large vision models or b) manually define 3D bounding regions within the scene representation. While the former is sensitive to suboptimal click locations and scales poorly with the number of objects, the latter demands significant domain expertise and high time complexity. Moreover,  existing approaches may rely on an additional NeRF model to assist with the subsequent multi-view mask propagation, which causes high computational costs and system complexity. 
2) \textbf{Limitations in inpainting and reconstruction.}
On the one hand, prominent pipelines such as NeRFiller and SPIn-NeRF require training an extra NeRF for depth estimation, followed by depth inpainting and another NeRF training for scene reconstruction. This typically takes over 100 minutes for the inpainting process alone, severely limiting scalability and practical utility for real-world applications.
On the other hand, these pipelines are usually tightly coupled with specific NeRF-based models, which also limits their generalizability.

In response to these challenges, we propose a more efficient and user-friendly pipeline (see Figure~\ref{fig:intro}) that improves both multi-view object mask generation and inpainting \& reconstruction, as illustrated in Figure~\ref{fig:comparision}. In the mask generation stage, our method introduces a more intuitive user interaction for obtaining the initial object mask on the source view: users can freely select arbitrary regions around the target object and then annotate a few key areas to preserve. This strategy improves both annotation accuracy and efficiency, particularly when dealing with multiple objects. For the subsequent multi-view mask generation, our method differs from prior approaches by eliminating the reliance on camera pose information, thereby avoiding potential errors introduced by pose noise. Additionally, it requires no extra model training. Instead, we leverage a lightweight HoMM module to directly produce high-quality multi-view object masks with enhanced efficiency.

In the inpainting and reconstruction stage, our pipeline further improves efficiency by bypassing the need to train an additional NeRF for depth maps and by avoiding inpainting on every single view. Rather than processing all views, we perform inpainting only on key source views which then propagate the results to other views using homography-based mapping, guided by the previously generated multi-view masks. Our method offers significantly improved efficiency while maintaining visual coherence. Meanwhile, in contrast to existing methods that are inherently bound to specific NeRF-based models, our pipeline can be seamlessly integrated into a variety of radiance field frameworks, including NeRF and 3D Gaussian Splatting. Comprehensive experiments demonstrate that our approach delivers competitive or superior quantitative and qualitative results, while reducing the runtime to just one-fifth of that required by state-of-the-art baselines, as seen in Fig. \ref{fig:comparision} (c).

Additionally, existing 3D multi-view removal datasets~\cite{mirzaei2023spin,weber2024nerfiller} lack diverse object samples necessary for tackling multi-object removal challenges. These datasets also have limited viewpoint variation, restricting the effective evaluation of multi-view consistency. In this regard, we introduce a comprehensive 3D multi-object multi-view removal dataset composed of real-world scenes. Extensive experiments on the new bench approach demonstrate its practicality and usefulness in real-world open scenarios. 
In both previous and new datasets, our method consistently achieves state-of-the-art performance, which further verifies the effectiveness of our method.
\noindent The main contributions of our work are as follows:
\vspace{-6pt}
\begin{itemize}
    \item We present HOMER, a plug-and-play homography-based object removal method compatible with any radiance field models, including both NeRF and 3D-Gaussian Splatting. 
    \item We develop an efficient and user-friendly multi-view mask generation framework that introduces intuitive region-based interaction in the source view and eliminates reliance on camera poses or additional model training, enabled by our lightweight HoMM module.   
    \item We collect a new 3D multi-object removal dataset with greater object diversity and wider viewpoint variation than existing datasets. Our method achieves state-of-the-art results on both public and our introduced datasets.
\end{itemize}

\section{Related Work}
\label{sec:related}

\subsection{Multi-View 3D Scene Understanding}

3D scene understanding methods can typically be divided into three categories~\cite{3DSurvey}: point-based methods~\cite{PointNet,pointnet++,ptv2,ptv3,DSPoint}, voxel-based methods~\cite{voxnet,choy20194d,pbnet}, and multi-view methods~\cite{bpnet,yang20232d}. The rapid development of visual language models~\cite{radford2021learning,rombach2022high,yang2024binding,yang2024neurobind} and foundation models~\cite{Transformer,dosovitskiy2020image,mamba} has empowered a useful foundation for downstream tasks~\cite{iQuery,DepthCLIP,zeng2024wordepth,zeng2024rsa}, particularly leading to a surge in methods that render 3D objects into 2D multi-view images for feature extraction or prediction. Specifically, some methods~\cite{ZhangGZLM0QG022,zhu2023pointclip,diffclip,zhao2025open} project the 3D object into multi-view images and leverage the CLIP~\cite{radford2021learning} or Diffusion~\cite{rombach2022high,ho2020denoising,zeng2024priordiffusion} to get the final recognition result. Moreover, methods~\cite{Genova21,bpnet,yang20232d} extract the multi-view images from videos of 3D scenes and utilize the 2D network to achieve the segmentation prediction. In light of this, our work combines the multi-view images and pre-trained 2D visual models~\cite{sun2021loftr,suvorov2022resolution} to attain the 3D scene removal without training labels. 

\vspace{-2pt}
\subsection{Object Removal in 2D Domain}

Most 2D object removal approaches~\cite{avrahami2023blended,avrahami2022blended,lugmayr2022repaint,nichol2021glide,xie2023smartbrush} erase the objects from RGB images or videos and engage the diffusion model~\cite{rombach2022high,ho2020denoising} to inpaint the erased images/videos. 
Several more challenging and highly relevant tasks are currently being explored, including large hole filling~\cite{li2022mat,suvorov2022resolution} and pluralistic inpainting~\cite{zheng2019pluralistic,wan2021high}. Moreover, Context-Encoders~\cite{pathak2016context} combine adversarial training with the encoder-decoder structure into this task. Furthermore, recurrent~\cite{li2020recurrent,zhang2018semantic}, gated convolution~\cite{yu2019free}, and multi-stage~\cite{ren2019structureflow} methods are proposed to enhance the effectiveness of object removal. More recent advancements, such as LaMa~\cite{suvorov2022resolution}, are designed to tackle the challenges associated with large missing regions and high-resolution images. Specifically, LaMa leverages fast Fourier convolutions to effectively capture global context from the initial stages of processing and incorporates a high receptive field perceptual loss and a strategic approach to mask generation during training. Since our goal is to blend the inpainted area with the surrounding background without the necessity of text prompts, our pipeline employs LaMa, which is better suited for achieving the desired level of coherence in object removal tasks. 
\vspace{-10pt}
\subsection{Object Removal in 3D Domain}
Current 3D object removal works~\cite{mirzaei2023spin,weder2023removing,mirzaei2023reference,chen2024mvip,viewconsistentobjectremovalradiance} mainly focus on alleviating inconsistencies in NeRF scene inpainting. Specifically, Remove-NeRF~\cite{weder2023removing} lightens inconsistencies using the inpainting confidence based view selection procedure; SPIn-NeRF~\cite{mirzaei2023spin} and InNeRF360~\cite{wang2024innerf360} employ the perceptual loss within inpainted regions to reduce inconsistencies between different views;  GScream~\cite{wang2024gscream} introduces Gaussian splatting as a method to bolster geometric consistency across the edited scenes. Also, NeRFiller~\cite{weber2024nerfiller} completes scenes and  iteratively utilizes the 2$\times$2 grid behavior of 2D inpainting models to 3D consistency. Additionally, Clutter-DR~\cite{wei2023clutter} employs 3D segmentation alongside instance-level area-sensitive losses to enhance the clarity and visual appeal of the inpainted scenes. 

Although these methods have achieved improvements, they are still limited in practical multi-object removal. First, the interaction for determining the objects to be removed in these methods is unfriendly and laborious. E.g., SOTAs~\cite{mirzaei2023spin,weder2023removing,mirzaei2023reference,chen2024mvip,maldnerf} either require multiple or iterative interactions to remove multiple objects, or they directly remove all objects of predefined categories in all 3D scenes~\cite{wei2023clutter}. In contrast, our method offers a more friendly pipeline to remove multiple objects merely in one single interaction. Furthermore, SOTAs~\cite{mirzaei2023spin,weder2023removing} demonstrate low efficiency in multi-view mask generation and matching. Especially, Remove-NeRF~\cite{weder2023removing} involves manually annotating a 3D box around the object using MeshLab, which is also complex and time-consuming for users. On the contrary, our method simplifies this process and enables the easy generation and matching of well-formed multi-view masks.

\begin{figure*}[ht]
\centering
\includegraphics[width=0.99\textwidth]{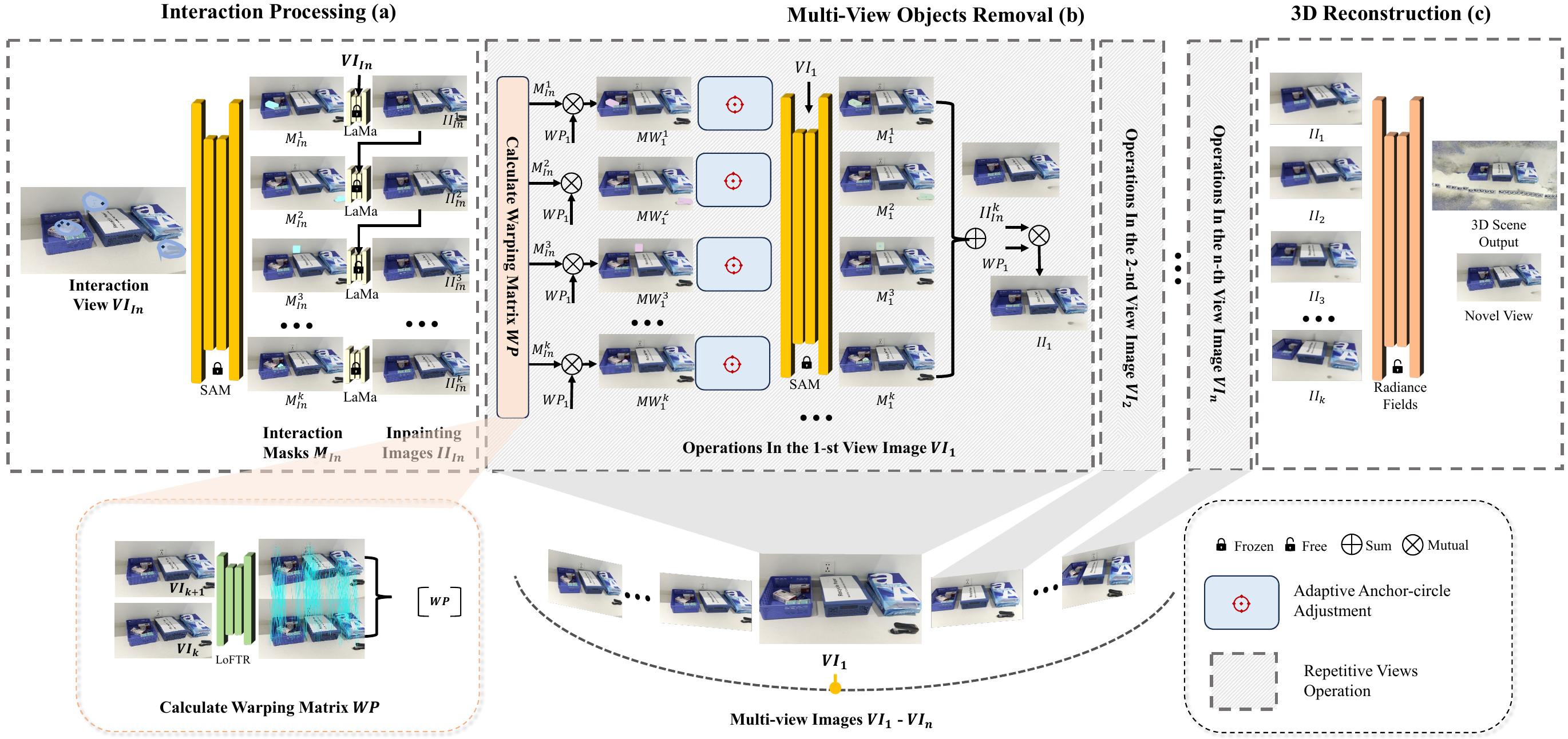} 
\caption{
  Network Architecture. Our proposed HOMER consists of three main stages: (a) Interaction Processing, (b) Multi-View Object Removal, and (c) 3D Reconstruction. 
  (a) In the interaction stage, the user selects a source view and specifies the regions to be removed and preserved. Binary masks are generated and sequentially inpainted by LaMa to produce the updated source image. 
  (b) In the multi-view removal stage, we compute homography matrices and warp the initial masks to other views to generate coarse masks. An adaptive anchor circle adjustment module is then used to generate final object masks for each view. Refined masks and inpainted results are propagated from the key views to other views with homography-based mapping.
  (c) The inpainted multi-view images and corresponding camera poses are fed into a radiance field model to reconstruct the 3D scene with objects removed.}
\label{fig:net}
\end{figure*}

\section{Proposed Method}
\label{sec:method}

\subsection{Overview}
Fig.~\ref{fig:net} presents an overview of our proposed pipeline, which consists of three main components: a) Interaction Processing, b) Multi-View Object Removal, and c) 3D Reconstruction. Multi-view images and corresponding camera poses are adopted as the network inputs. We begin by selecting an arbitrary source view for user interaction, where the user specifies the region of objects to be removed and the areas to be preserved through point prompts. These prompts will then be used as inputs to a pre-trained segmentation model ~\cite{sam,sam2} for producing a binary mask of the objects to be removed, which is then combined with a pre-trained inpainting model ~\cite{suvorov2022resolution} to inpaint the masked-out area from the selected source view.

To propagate mask correspondences across views, we leverage LoFTR~\cite{sun2021loftr} to compute dense keypoint matches between adjacent views and estimate homography matrices. These matrices are used to warp the initial binary mask from the source view to neighboring views, producing coarse masks. Based on these coarse masks, we introduce an adaptive anchor circle adjustment algorithm that generates point prompts to be input for the segmentation model in the neighboring view. These prompts guide the extraction of refined object masks in each target view.
Moreover, we utilize the previously estimated homography matrices and refined masks to propagate inpainting results. Starting from the source view, we warp the inpainted area to other views and identify a set of key views for direct inpainting using LaMa to mitigate error accumulation. Finally, the inpainted multi-view images and their corresponding camera poses are used to train a radiance field with objects removed.

\subsection{Interaction Processing}
\noindent\textbf{Interaction.} As illustrated in Fig. \ref{fig:net}(a), the interaction process utilizes user-provided prompts to define arbitrary regions for object removal, along with point prompts to specify the objects or areas to be preserved. Let $P_{fg} = \{p_{fg}^j\}_{j=1}^{N_{fg}}$ represent the set of $N_{fg}$ foreground points indicating pixels within the target objects to be removed, and $P_{bg} = \{p_{bg}^l\}_{l=1}^{N_{bg}}$ represent the set of $N_{bg}$ background points identifying pixels in areas to be preserved. Each point $p = (x, y)$ is a coordinate within the image dimensions ($0 \le x < W, 0 \le y < H$). The complete set of prompts is $P = P_{fg} \cup P_{bg}$. These prompts need only be placed on a single, user-selected view ($VI_{In}$) where the boundaries of the target object(s) are clearly discernible.

\noindent\textbf{Segmentation and Inpainting of Interaction View.}
Given the interactive view image $VI_{In}$ and the set of point prompts $P$, we employ a pre-trained segmentation model \cite{sam2, sam} to generate binary masks for the $K$ target objects. This yields the set of interaction masks $M_{In} = \{ M_{IN}^i \in \{0, 1\}^{W \times H} \}_{i=1}^{K}$, where $M_{IN}^i$ is the binary mask for the $i$-th object. Within each mask, a value of 1 indicates a pixel belonging to the object designated for removal, while 0 indicates a pixel to be retained.

As depicted in Fig.~\ref{fig:net}(a), these masks $M_{In}$ guide a sequential object removal and inpainting process using a pre-trained inpainting model (e.g. LaMa). Starting with the initial interactive view, $II_{in}^{0} = VI_{In}$, for each object $i$ we use its corresponding mask $M_{In}^{i}$ to guide the inpainting model, operating on the result from the previous step:
\begin{equation}
\label{eq:inpainting_step}
II_{in}^{i} = \text{LaMa}(II_{in}^{i-1}, M_{In}^{i}), \quad \text{for } i = 1, \dots, K,
\end{equation}

where, $\text{LaMa}(Image, Mask)$ denotes the operation where the LaMa model inpaints the regions specified by the mask (where mask value is 1) in the input image. After $K$ steps, the resulting image $II_{in}^{K}$ contains the interactive view with all $K$ specified objects removed and the corresponding areas inpainted. Both the final inpainted image $II_{in}^{K}$ and the set of masks $M_{In}$ serve as key inputs for our subsequent multi-view object removal process.
Our image inpainting can be performed in two main ways: sequentially inpainting each mask, or inpainting all masks at once after merging them. We provide implementations for both approaches, which can be selected based on the actual application scenario. The sequential inpainting method is generally suitable for situations with many objects and uniform colors, while the single-step inpainting method is more appropriate for scenarios with richer colors or time constraints.

\subsection{Multi-View Object Removal}
We propose a Homography-based Mask Matching (HoMM) Module for multi-view object removal. Given a set of multi-view images \( VI = \{ VI_j \in \mathbb{R}^{W \times H \times 3} \}_{j=1}^{N_v} \), where \( VI_j \) denotes the \( j \)-th view and \( N_v \) represents the total number of views. Our method achieves object removal by accurately mapping object masks across views and performing subsequent warping-based inpainting.

Prior to processing, we apply RGB image preprocessing and extract dense correspondences using LoFTR~\cite{sun2021loftr}, which provides viewpoint similarity measurements and matching keypoints. To reduce the computational cost, we consider only adjacent view pairs rather than all \( N_v \cdot (N_v - 1) \) possible combinations, effectively reducing the total number of required pairs to \( N_v - 1 \).

We compute the homography matrix using obtained matching points for each selected image pair. The steps for each view are identical, with the mask $M_{In}$ and image $II_{in}$ from the previous view serving as the matching information for the next view. For simplicity, we describe the process for a single view as shown in Fig.~\ref{fig:net}(b). The operations for each view contain three main parts: 1) generating coarse masks by warping the previous view’s masks with the homography matrix, 2) refining the coarse mask through our adaptive anchor circle adjustment module, and 3) performing warping-based inpainting using the inpainted result from the previous view. We detail these three components as follows.

\noindent\textbf{Homography Matrix for Mask Warping.} 
Homography matrix represents a projective transformation between two planes. In the context of planar surfaces, this transformation can be represented by a 3×3 matrix $H$ that maps points from one plane to another while preserving collinearity and cross-ratio properties. Given a point in homogeneous coordinates, the homography transformation can be:
$$ \begin{bmatrix} x' \\ y' \\ w' \end{bmatrix} = 
\begin{bmatrix} 
h_{11} & h_{12} & h_{13} \\
h_{21} & h_{22} & h_{23} \\
h_{31} & h_{32} & h_{33}
\end{bmatrix}
\begin{bmatrix} x \\ y \\ 1 \end{bmatrix}. $$
Each matched point pair $(p_i, p'_i)$ provides two constraint equations for computing the homography matrix. While theoretically four point correspondences are sufficient to determine the homography matrix due to its eight degrees of freedom, they would be susceptible to noise and matching errors. We iteratively sample minimal sets of four correspondences to compute candidate homography matrices. For each iteration, a candidate matrix is computed and evaluated against all matching points using the reprojection error:
$$ \varepsilon = \sum_{i=1}^{n} ||p'_i - H \cdot p_i||^2. $$
This identifies inliers (points with error below a threshold) and selects the homography matrix that maximizes the number of inliers, so as to find an optimum best fiting the majority of correspondences.

Given a binary mask $M_i$ at view $i$ and the computed homography matrix $H_{i,i+1}$ between consecutive images $i$ and $i+1$, the warping process transforms $M_i$ to align with image $i+1$, resulting in the warped mask $MW_{i+1}$. For each pixel position in the target image, we compute its corresponding position in the source image using the inverse homography transformation:
$$ MW_{i+1}(x, y) = MW_i(H_{i,i+1}^{-1}\begin{bmatrix} x \\ y \\ 1 \end{bmatrix}). $$
This transformation maintains the topology of the mask region while adjusting its geometric shape according to the perspective change between multi-view images.

\noindent\textbf{Adaptive Anchor Circle Adjustment.} 
Through the homography matrices, we can obtain the warping mask $MW_i^k, MW_i^k \in  \mathbb{R}^{ W \times H}$ for the $k$-th object to be removed in the $i$-th view, i.e., our coarse masks.  
However, the accuracy of the estimated homography matrices inherently depends on the precision of the underlying point correspondences. In scenarios with large viewpoint variations, such as those captured under sparse views, homography estimation becomes less reliable, which will lead to noticeable warping errors. Moreover, these errors tend to accumulate progressively across multiple warping steps. To tackle this issue, we introduce an adaptive anchor circle adjustment strategy in conjunction with the pre-trained segmentation model~\cite{sam2,sam2} to refine the initial masks. 
The center of mass point $C_i^k$ of the warping mask $MW_i^k$ serves as the initial center of the anchor circle with initial radius $r$. We sample points from the initial anchor circle to serve as the point prompts for the $i$-th view image $VI_i$. Furthermore, we combine the point prompts and $VI_i$ as the input to the segmentation model to obtain the initial object mask $M_i^k$. We treat $r$ as an optimization variable for gradient descent and use the IoU and Shape Context Distance between $M_i^k$ and $MW_i^k$ as the optimization constraint. The IoU metric ensures overall mask alignment, while the Shape Context Distance captures fine-grained shape similarities. This dual-constraint optimization can be formulated as:
\begin{equation}
\mathcal{L} = \alpha(1 - \text{IoU}(M_i^k, MW_i^k)) + \beta \cdot \text{SC}(M_i^k, MW_i^k),
\end{equation}
where $\text{SC}(\cdot)$ represents the Shape Context Distance between two masks, and $\alpha, \beta$ are weighting coefficients balancing the contribution of each term. We iteratively find the optimal value of $r$ by maximizing the equation. As a result, we obtain the refined final masks \( M_i^k \) across all viewpoints.

\noindent\textbf{Warping-Based Inpainting.} After obtaining the refined masks \( M_i^k \) for each view, we perform warping-based inpainting to propagate the inpainted content across views. Starting from the interactive source view \( II_{\text{in}} \), we warp the pixel content within the masked region to the next view using the homography matrix \( H_{i, i+1} \). For each object \( k \) to be removed, the warped RGB content \( \hat{I}_{i+1}^k \) is computed as:
\begin{equation}
\hat{I}_{i+1}^k(x, y) = II_i^k(H_{i+1,i}^{-1} \cdot [x, y, 1]^T),
\end{equation}
where $(x, y)$ in $M_{i+1}^k$, and \( M_{i+1}^k \) denotes the refined mask of object \( k \) in view \( i+1 \). The warped content is then combined with \( M_{i+1}^k \) by computing the intersection. If large unfilled regions remain in the mask (e.g., occluded areas), we apply LaMa~\cite{suvorov2022resolution} to selectively inpaint those regions:
\begin{equation}
II_{i+1}^k = \text{LaMa}( \hat{I}_{i+1}^k, M_{\text{empty}} ),
\end{equation}
where $M_{\text{empty}} = M_{i+1}^k \setminus \text{mask}(\hat{I}_{i+1}^k)$,
and $\text{mask}(\hat{I}_{i+1}^k)$ indicates the valid pixels in the warped image.

To avoid the accumulation of warping errors, especially under large viewpoint variations, we perform inpainting every n views (e.g., \( n=10 \)) instead of continuously propagating content across all frames. Finally, the complete set of inpainted images is denoted as:
${II_j}_{j=1}^{N_v}$, where $II_j = \text{inpainted image of view } j \text{ with objects removed}$.

\subsection{3D Reconstruction}
As shown in Fig. \ref{fig:net}(c), we utilize radiance fields to refine the multi-view consistency of inpainting results and reconstruct the 3D scene. NeRF~\cite{mildenhall2021nerf} models a scene by learning a mapping from 3D coordinates and views directions to color and density using one multi-layer perceptron (MLP), $F_\theta$. For a given ray $r$, the estimated color $\hat{C}(r)$ is computed via volumetric rendering:
\begin{equation}
\hat{C}(r)_{NeRF} = \sum_{i=1}^{K} T_i (1 - \exp(-\sigma_i \delta_i)) c_i,
\end{equation}
where $c_i$ and $\sigma_i$ are the color and density at the $i$-th sample along the ray, $\delta_i = t_{i+1} - t_i$ represents the distance between consecutive samples, and $T_i = \exp\left(-\sum_{j=1}^{i-1} \sigma_j \delta_j\right)$ represents the accumulated transmittance along the ray. 

The model is trained by minimizing the loss between the rendered images and the ground-truth images from the training set:
\begin{equation}
L_{NeRF} = \sum_{r \in \mathcal{R}} \|\hat{C}(r)_{NeRF}  - C_{GT}(r)\|^2,
\end{equation}
where $\mathcal{R}$ is a batch of rays sampled from the training views, and $C_{GT}(r)$ denotes the ground-truth color for ray $r$. 

In addition to NeRF, we also support 3D Gaussian Splatting (3D-GS)~\cite{3dgs}. 3D-GS models the scene as a collection of anisotropic 3D Gaussians, each parameterized by its center \( \mu_i \), covariance \( \Sigma_i \), opacity \( \alpha_i \), and color \( c_i \). The color \( \hat{C}(r) \) observed along a ray is computed through alpha compositing of Gaussians sorted by depth:
\begin{equation}
\hat{C}(r)_{GS}  = \sum_{i} \alpha_i \cdot c_i \cdot \prod_{j < i}(1 - \alpha_j),
\end{equation}
and optimized with the same reconstruction loss:
\begin{equation}
L_{GS}  = \sum_{r \in \mathcal{R}} \left\| \hat{C}(r)_{GS}  - C_{\text{GT}}(r) \right\|^2.
\end{equation}
3D-GS provides a rendering-efficient alternative to NeRF, and our framework enables flexible deployment across diverse radiance field backbones.

\begin{figure*}[ht]
    \centering
    \includegraphics[width=\linewidth]{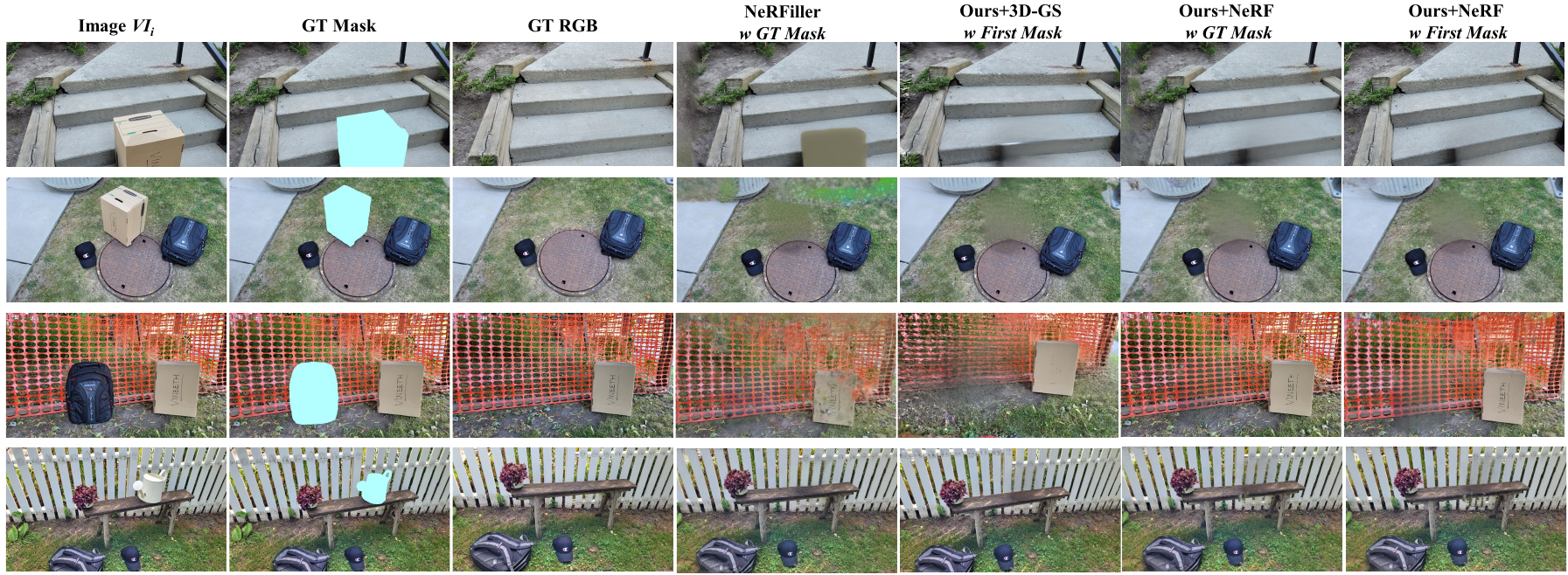}
    \caption{{\bf NeRF renderings for our dataset.} $w$ stands for $with$. We show the rendering of the first test pose for 4 sampled scenes in our dataset. We compare our method to the NeRFiller results using DINO masks, as well as using our generated masks on the second row, and our method on the third row.} 
    \label{fig:oursr}
\end{figure*}
\section{Experiments}
\label{sec:experiments}

\subsection{Experiment Setting}
All of our evaluations are conducted on a single NVIDIA GeForce RTX 3090 GPU with 24GB VRAM. 
\\
\textbf{Datasets.} We validate the quality, effectiveness, and applicability of our method on two datasets: the SPIn-NeRF dataset~\cite{mirzaei2023spin} and our proposed multi-object open scenes dataset. The SPIn-NeRF dataset is currently one of the most popular datasets and is used by state-of-the-art methods. It includes ground truth for scenes after object removal. However, it contains only 10 scenes, each with approximately 1 to 3 objects, and focuses on the removal of a single masked object per scene. In contrast, our proposed multi-object open scenes dataset comprises 24 scenes, each featuring more than five removable objects. All images in our dataset are extracted from video frames, with each video uniformly sampled to obtain 75 or 100 frames. The videos were recorded in 4K resolution using a Samsung Galaxy S22 Ultra and subsequently downsampled to 1920 × 1080 for further processing. Our dataset encompasses a diverse range of indoor and outdoor environments, including both simple and complex scenes, making it more representative of real-world scenarios. Additionally, it introduces greater viewpoint variation, increasing the complexity and robustness of the dataset for multi-object removal tasks. Our dataset will be publicly available upon acceptance of the paper.

\noindent\textbf{Evaluation Metrics.} Following recent works ~\cite{mirzaei2023spin,wang2024innerf360,weber2024nerfiller}, we adopt standardized metrics PSNR, SSIM, and LPIPS to evaluate the quality of the rendered images after object removal. In addition, we measure the efficiency in terms of runtime. We also report the Fréchet Inception Distance (FID) to assess the perceptual similarity between the rendered images and the ground truth. Additionally, for our dataset, we evaluate the average time for the model to train from start to convergence, including the inpainting process.

\noindent\textbf{Camera Pose Estimation.}
We employ COLMAP~\cite{colmap1,colmap2} to obtain camera poses only for reconstruction purposes. COLMAP is an open-source Structure-from-Motion (SfM) system that implements a robust incremental reconstruction pipeline. This system enables the recovery of camera parameters and sparse 3D scene geometry from uncalibrated image collections.

\noindent\textbf{Radiance Fields Reconstruction.}
We utilize NeRFStudio~\cite{nerfstudio} for NeRF reconstruction and 3D Gaussian Splatting reconstruction. NeRFStudio is an open-source framework that provides standardized implementations of various NeRF methods, enabling efficient training and evaluation of novel view synthesis techniques. For all our experiments, we employ NeRFStudio's default nerfactor model~\cite{nerfactor} and default splatfacto model. We conduct all reconstructions using the default configuration without parameter tuning. 

\subsection{Results}
\noindent\textbf{Multi-view Masks Matching.} We evaluate multi-view mask matching on our dataset. As NeRFiller does not provide object masks by default, we adopt SPIn-NeRF pipeline and our method for mask generation comparison.
Following SPIn-NeRF, we select object points in the first frame, and the pre-trained SAM2 model \cite{sam2} is applied as our segmentation model for initial mask generation and subsequent mask refinement. These masks serve as input to the DINO model, which propagates the segmentation results to subsequent frames. Our method starts with user interaction to create the masks from any selected viewpoint. Then we propagate the initial masks to the remaining views using our HoMM module. The two resulting sets of multi-view masks are provided to NeRFiller for reconstruction and object removal.

As shown in Tab.~\ref{tab:evod} and Fig.~\ref{fig:oursr}, the mask quality and accuracy obtained through SPIn-NeRF is inferior to our HoMM module. Our analysis reveals two primary factors contributing to this quality disparity. First, SPIn-NeRF's point-based selection approach often results in incomplete object coverage during the initial selection phase, leading to partial object omission. Second, when DINO generates masks for alternative viewpoints, it introduces substantial distortions to the object's shape and boundaries. Furthermore, the mask propagation process in DINO exhibits inherent tracking instability, leading to incomplete object removal outcomes.
\begin{figure}[ht]
    \centering
    \includegraphics[width=0.80\linewidth]{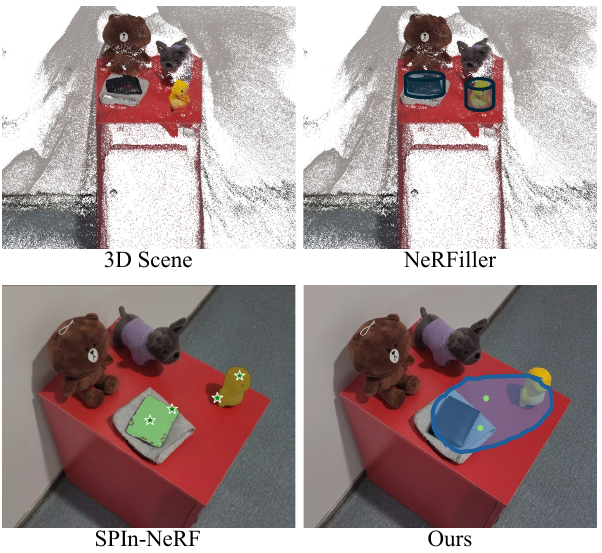}
    \caption{Interaction Comparison} 
    \label{interaction}
\end{figure}

\begin{table}[ht]
\centering
\resizebox{0.40\textwidth}{!}{
\begin{tabular}{l|c|c|c}
\toprule
Methods   &  PSNR $\uparrow$&  SSIM$\uparrow$ &  LPIPS$\downarrow$ \\ \hline
NeRFiller + SPIn-NeRF &27.42&0.85&0.14\\
 NeRFiller + Ours    &\textbf{30.38}&\textbf{0.86}&\textbf{0.11} \\ 
\bottomrule
\end{tabular}
}
\caption{Evaluation on Our Dataset}
\label{tab:evod}
\end{table}

\begin{figure*}[ht]
    \centering
    \includegraphics[width=0.97\linewidth]{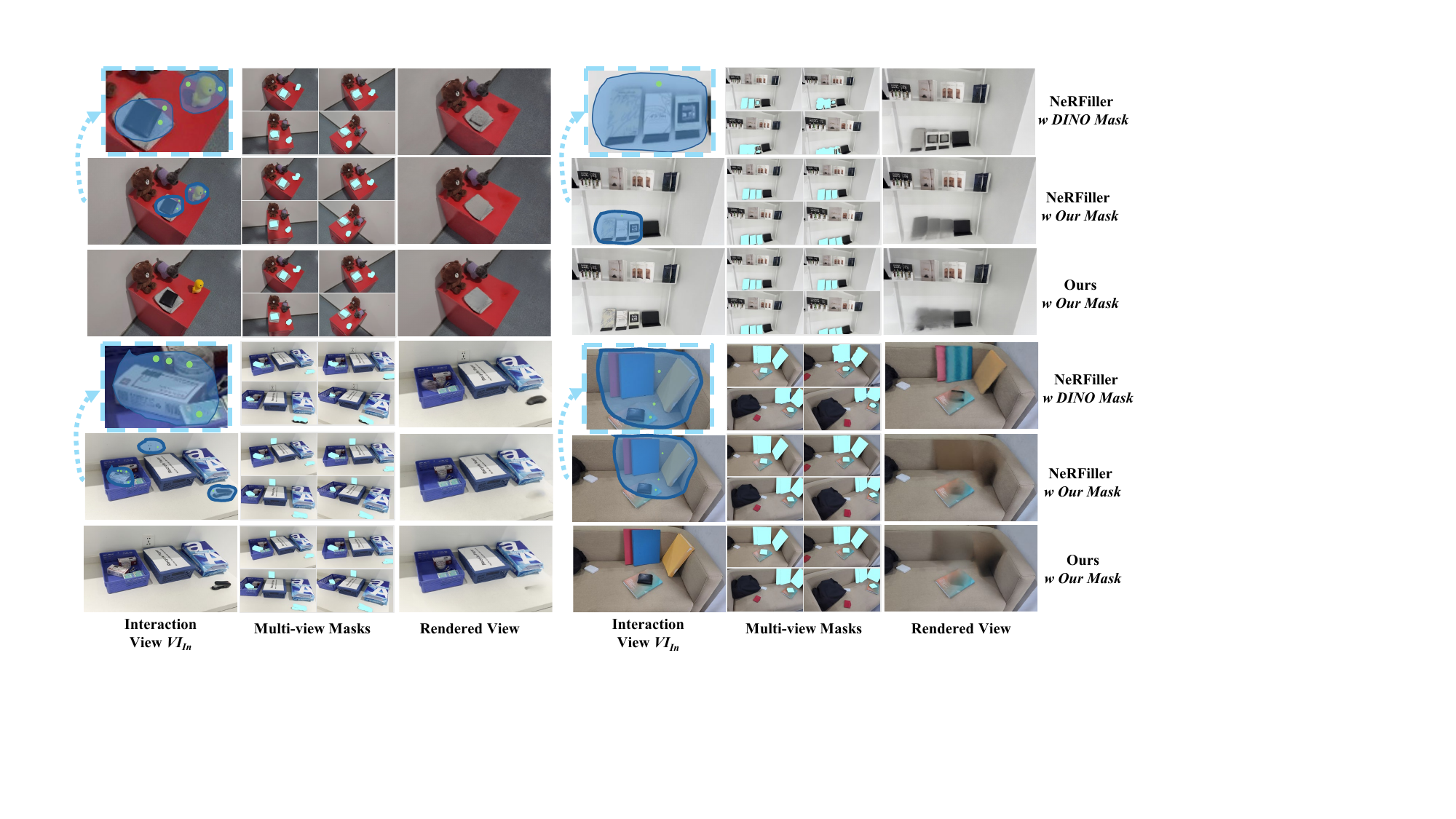}
    \caption{{\bf NeRF renderings for SPIn-NeRF dataset.} $w$ stands for $with$ and GT denotes ground-truth. We show the rendering of the first test pose for 4 sampled scenes in the SPIn-NeRF dataset. We compare our method to the NeRFiller results using ground truth masks, as well as using our generated masks on the third row.} 
    \label{fig:spinr}
\end{figure*}

\noindent\textbf{Multi-view Inpainting.}
The SPIn-NeRF dataset with ground truth is utilized to evaluate the quality and consistency of multi-view inpainting. For SPIn-NeRF, our result includes the values reported in the paper~\cite{mirzaei2023spin}. For NeRFiller, we follow the instructions provided in the paper to use reference inpainting and render the test view images for comparison with the ground truth. 

For our proposed method, we adopt two configurations: 1) leveraging the accurate multi-view masks produced by SPIn-NeRF to perform warping-based inpainting, and 2) using only the mask from the first view and applying our HoMM module to generate the corresponding masks for the remaining views. In both settings, we render images from novel test views and compute LPIPS and FID metrics against the ground truth.
We apply both configurations to NeRF-based reconstructions, while the more challenging single-view mask setup is used for 3D Gaussian Splatting (3D-GS). As presented in Table~\ref{tab:evsp} and Fig.~\ref{fig:spinr}, our method achieves LPIPS scores comparable to the other two baselines, while significantly outperforming them in terms of FID.
As reported in Table~\ref{tab:evsp}, one scene in the 3D Gaussian Splatting (3D-GS) experiments shows anomalously high FID and LPIPS values, which we attribute to suboptimal reconstruction quality produced by the default splatfacto model on that specific scene. This limitation could potentially be mitigated by employing the splatfacto-big model in NeRFStudio, which offers improved reconstruction fidelity at the cost of increased memory consumption. After excluding this outlier, the combination of our method with 3D-GS achieves the best FID performance.

\begin{table}[ht]
\centering
\resizebox{0.27\textwidth}{!}{
\begin{tabular}{l|c|c}
\toprule
Methods   &  LPIPS$\downarrow$ &FID$\downarrow$ \\ \hline
SPIn-NeRF  &0.47&156.64   \\
SPIn-NeRF$^{\dagger}$ & \textbf{0.46} & 147.31\\
NeRFiller  &0.53     &69.40 \\  
Ours+3D-GS$^{\alpha}$ & 0.57$^*$ & 96.06$^*$ \\  
Ours+NeRF     & 0.52&    51.78 \\ 
Ours+NeRF$^{\alpha}$ &0.56 & \textbf{50.35} \\
\bottomrule
\end{tabular}
}
\caption{
\textbf{Evaluation on SPIn-NeRF Data.} SPIn-NeRF$^{\dagger}$ denotes SPIn-NeRF with refined RGB inputs. $^{\alpha}$ indicates methods  using only the 1st ground-truth mask.
$^*$ There is one outlier in GS experiments. After excluding this outlier, FID is 49.99 and LPIPS is 0.55.
}

\label{tab:evsp}
\end{table}

\noindent\textbf{Efficiency.}
The comparative evaluation of interaction time and removal time highlights the efficiency of our method in handling multi-object scenarios. As shown in Fig. \ref{interaction} and Tab. \ref{ea}, our method offers more usability through its streamlined and intuitive interface. When handling two objects simultaneously, our approach requires only 3.4 seconds, improving by 24.4\%  over SPIn-NeRF and by 99.4\% reduction compared to NeRFiller. This efficiency gap widens further in three-object scenarios, where our method completes processing in 3.6 seconds, achieving a 29.4\% improvement over SPIn-NeRF and a 99.6\% reduction compared to NeRFiller. In addition, our method maintains near-constant time complexity with increasing object count. The processing time increases by only 0.4 seconds when moving from one to two objects, and by 0.2 seconds when advancing to three objects. In contrast, other methods show obvious increases in interaction time as the object number grows. 

The removal time metric further underscores our method's efficiency, requiring only 20 minutes compared to 100 minutes for SPIn-NeRF and NeRFiller, and 115 minutes for SPIn-NeRF with refined RGB. This represents an 80\% reduction in removal time versus the standard implementations and an 82.6\% improvement over SPIn-NeRF with refined RGB.

These results demonstrate that our method not only achieves faster processing times but also maintains consistent performance across varying levels of scene complexity. Such characteristics make our approach particularly suitable for real-world applications where complex, multi-object scenes are common.

\begin{table}[ht]
\centering
\resizebox{0.45\textwidth}{!}{
\begin{tabular}{l|ccc|c}
\toprule
Methods & \multicolumn{3}{c|}{Interaction Time (sec.) $\downarrow$} & Remove  \\ \cline{2-4}
        & One Obj.& Two Obj. & Three Obj. & Time (min.) $\downarrow$\\ \hline
SPIn-NeRF              & 3.1  & 4.5  & 5.1  & 100  \\
SPIn-NeRF$^{\dagger}$ & 3.1  & 4.5  & 5.1  & 115  \\
NeRFiller              & 305.5  & 612.5 & 917.5 & 100 \\  
Ours+NeRF & \textbf{3.0} & \textbf{3.4}  & \textbf{3.6} & 20\\ 
Ours+3D-GS & \textbf{3.0} & \textbf{3.4 }  & \textbf{3.6 } & \textbf{15}\\ 
\bottomrule
\end{tabular}
}
\caption{Comparison on the Time Cost. Obj. stands for the object(s).}
\label{ea}
\end{table}

\section{Conclusion}
In this paper, we present HOMER, a novel interaction-driven framework for multi-object removal in 3D scenes. Our method enables intuitive user interaction by allowing the specification of target objects and background regions from any arbitrary viewpoint. To address the inherent challenges of multi-view mask correspondence and propagation, we introduce a lightweight Homography-based Mask Matching (HoMM) module. This module computes homography matrices from keypoint correspondences and integrates an adaptive anchor circle adjustment strategy to ensure geometric consistency across views.
By decoupling the mask generation process from camera pose dependency and additional model training, our approach achieves both high usability and computational efficiency. Furthermore, our inpainting pipeline avoids the need for depth estimation or repeated NeRF optimization by operating directly on selected key views and propagating results using homography-based mapping. Extensive experiments on both public and newly introduced datasets demonstrate that HOMER achieves competitive perceptual quality while significantly reducing computational overhead. Our framework supports various radiance field models, such as NeRF and 3D Gaussian Splatting, and is well-suited for deployment in real-world scenarios.

\bibliographystyle{ACM-Reference-Format}
\bibliography{homer}
\end{document}